\documentclass[sigconf,screen]{acmart}
\renewcommand\footnotetextcopyrightpermission[1]{}
\usepackage{subcaption}
\usepackage{multirow}

\acmConference[MRAC '26]{4th International Workshop on Multimodal, Generative and Responsible Affective Computing}{November 10--14, 2026}{Rio de Janeiro, Brazil}

\begin{document}

\title{Beyond Ambiguous Visual Cues: Studying Physiological Disruptions and Cross-Modal Inconsistencies in Deepfake Videos
}

\author{Chenxi Yang}
\authornote{Equal Contribution}
\email{YCX-SPEIT@sjtu.edu.cn}
\affiliation{%
  \department{LIASD Laboratory}
  \institution{Shanghai Jiaotong University}
  \city{Shanghai}
  \country{China}
}
 \orcid{https://orcid.org/0009-0005-6209-6794}

\author{Yassine Ouzar}
\authornotemark[1]
\authornote{Corresponding author}
\email{yassine.ouzar@univ-paris8.fr}
\affiliation{%
  \department{LIASD Laboratory}
  \institution{University of Paris 8}
  \city{Saint-Denis}
  \country{France}
}

\orcid{https://orcid.org/0000-0002-7327-5895}

\author{Larbi Boubchir}
\email{larbi.boubchir@univ-paris8.fr}
\affiliation{%
  \department{LIASD Laboratory}
  \institution{University of Paris 8}
  \city{Saint-Denis}
  \country{France}
}

\orcid{https://orcid.org/0000-0002-5668-6801}

\renewcommand{\shortauthors}{Yang et al.}
\renewcommand{\shorttitle}{Studying Physiological Disruptions and Cross-Modal Inconsistencies in Deepfake Videos}

\begin{abstract}
Recent deepfake detection studies increasingly suggest remote photoplethysmography (rPPG) signals as an authenticity cue. However, existing benchmarks lack physiological ground truth, and current detectors underexplore the cross-level relationship between facial features and physiological dynamics, often relying on late fusion or rPPG features alone.
In this paper, we construct high-fidelity deepfake manipulations on established real rPPG datasets (COHFACE and UBFC-rPPG) to investigate how forgeries disrupt natural physiological signals and facial behavior at the same time. Building on this analysis, we propose a bidirectional co-attention fusion detector that jointly models rPPG and facial behavior tokens. This mechanism explicitly captures the cross-level dependencies between pulse dynamics and facial motion to learn a robust, joint authenticity representation.
Extensive experiments using a subject-disjoint 5-fold evaluation demonstrate the superiority of our approach. Achieving a 92.80\% AUC on constructed datasets using face swapping and 96.78\% AUC on motion transfer, our model outperforms both the rPPG-only single modality baseline and the best feature-level fusion methods. Furthermore, transfer-learning result of the fusion detector on Celeb-DF-v2 while keeping both feature extractors fixed achieves 91.20\% accuracy and 86.08\% AUC, which suggests applicability under target-domain adaptation.

\end{abstract}



% \ccsdesc[100]{Applied computing~Health informatics}



\maketitle

\section{Introduction}

Deepfake generation has rapidly improved the visual realism of facial videos, making manipulated content increasingly difficult to identify from appearance artifacts alone. Early deepfake detectors often relied on spatial artifacts~\cite{afchar2018mesonet,matern2019visual}, face-warping or blending boundaries~\cite{li2019warping,li2020facexray}, and frame-level behavioral inconsistencies such as abnormal eye blinking, head-pose inconsistency, and unnatural lip dynamics~\cite{jung2020deepvision,yang2019headpose,haliassos2021lips}. However, as generative models have evolved from GANs to high-fidelity StyleGAN and diffusion-based synthesis~\cite{goodfellow2014gan,karras2020stylegan2,rombach2022latentdiffusion}, and as face manipulation benchmarks continue to expose stronger and more realistic forgeries~\cite{rossler2019faceforensics,li2020celebdf}, visual cues can become ambiguous, dataset-dependent, or short-lived. This motivates detection methods that look beyond surface-level appearance and examine whether a generated face preserves intrinsic temporal properties of real human videos.

Physiological signals offer one such direction. Prior studies have suggested that synthetic facial videos may fail to preserve remote photoplethysmography (rPPG) traces, leading to the intuition that deepfakes may ``lack a heart''~\cite{dwivedi2026rppgnet}. rPPG captures pulse-related color variations from facial videos and is therefore semantically different from conventional visual artifacts. Nevertheless, current physiology-aware deepfake detection still faces two important limitations. First, widely used deepfake benchmarks such as FaceForensics++~\cite{rossler2019faceforensics} and Celeb-DF~\cite{li2020celebdf} were not designed with physiological ground-truth measurements, making it difficult to verify whether extracted rPPG features are physiologically meaningful or merely dataset-specific artifacts. Second, existing methods often use physiological signals through handcrafted rules, independent classifiers, or late fusion~\cite{ciftci2019fakecatcher,dwivedi2026rppgnet}, leaving the relationship between physiological dynamics and facial behavior underexplored.

\begin{figure*}[t]
 \includegraphics[width=\textwidth]{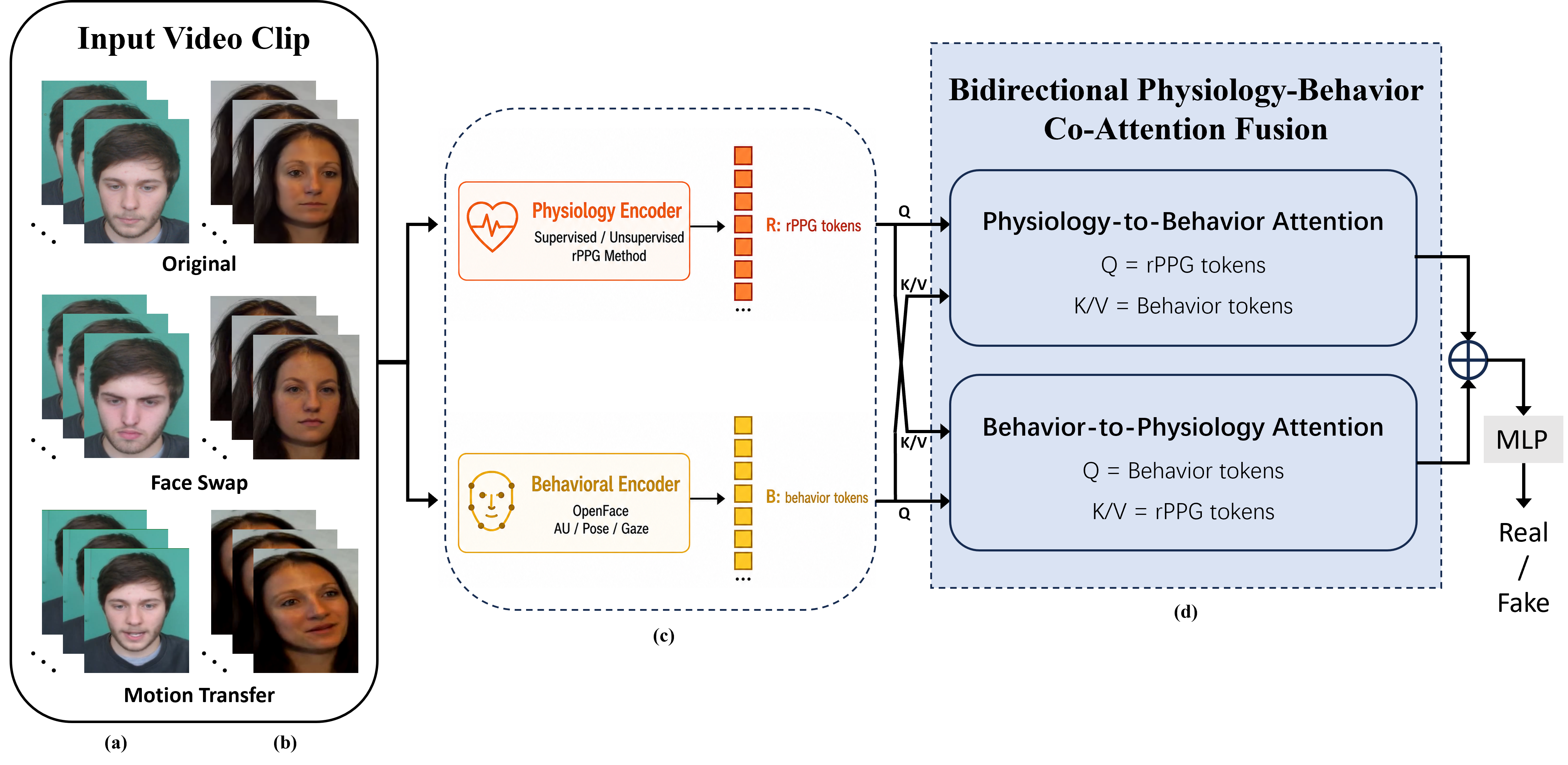}
  \caption{Overview}
  \vspace{0.3em}
  \begin{minipage}{0.96\textwidth}
  \small
 As illustrated in Figure~\ref{fig:overview}, we first construct face-swap and motion-transfer videos from rPPG datasets with synchronized physiological references, as detailed in Section~\ref{sec:dataset_construction}. The recovered rPPG signals are further analyzed in Section~\ref{sec:physio_analysis} to characterize how different manipulations affect physiological fidelity. For each $T$-frame clip, the two encoders then produce aligned rPPG tokens $R\in\mathbb{R}^{T\times96}$ and behavior tokens $B\in\mathbb{R}^{T\times26}$. Bidirectional cross-attention allows each modality to retrieve complementary temporal evidence from the other. After residual normalization and masked mean pooling, the two representations are concatenated and passed to an MLP for real--fake classification. The effectiveness of this fusion design is evaluated against single-modality and shallow-fusion baselines in Section~\ref{sec:fusion_analysis}.
  \end{minipage}
  \Description{The framework extracts temporally aligned rPPG and facial   behavior tokens and models their interactions through bidirectional   cross-attention before real--fake classification.}
  \label{fig:overview}
\end{figure*}

In this paper, we study deepfake detection through physiological disruptions and cross-level fusion between rPPG signals and facial behavior. To support this study, we construct high-fidelity deepfake manipulations on established real rPPG datasets, including COHFACE~\cite{heusch2017cohface} and UBFC-rPPG~\cite{bobbia2019ubfc}. This setting allows us to analyze manipulated videos whose real counterparts have physiological reference signals. Building on this analysis, we propose a bidirectional co-attention fusion detector that jointly models rPPG tokens and facial behavior tokens extracted from action units (AU), pose, and gaze, shown in Figure ~\ref{fig:overview}. Instead of treating rPPG as an isolated detection feature, we ask whether manipulated videos preserve the internal relationship between pulse-related dynamics and facial behavior. Blood volume changes induce subtle skin-color fluctuations that form the basis of rPPG estimation in pixels, whereas AU activations provide an interpretable representation of expression-related facial muscle dynamics at a macroscopic level. 

Our main contributions are summarized as follows:

\begin{itemize}
    \item 
    We introduce a physiology-grounded manipulations, applying multiple deepfake generation techniques, including face swapping and motion transfer, to real rPPG datasets with physiological reference signals. This design enables controlled analysis of how deepfake manipulations affect physiological facial dynamics.

    \item We provide a manipulation-wise analysis of physiological disruptions in deepfake videos. By comparing real and manipulated videos as well as analyzing feature changes within manipulated videos, we empirically validate the feasibility of using rPPG-related physiological cues for deepfake detection.

    \item We propose a bidirectional co-attention fusion detector that explicitly models cross-modal dependencies between rPPG dynamics and facial behavior features, including action units, pose, and gaze. Experiments show that the proposed fusion strategy outperforms single-modality baselines and shallow fusion methods such as concatenation and weighted fusion.
\end{itemize}

\section{Related Work}

\subsection{Deepfake Generation}
Deepfake generation can be grouped by which component of a video is manipulated: identity, facial dynamics, or the entire visual content.~\cite{pei2026deepfakeSurvey}.

In face swapping, the source identity is transferred onto a target video while the target motion and scene context are largely preserved. Early practical systems and non-adversarial pipelines include DeepFaceLab and segmentation-based face swapping~\cite{perov2020deepfacelab,nirkin2018facesegmentation}, while GAN-based methods such as FaceShifter, SimSwap, and FSGAN improve identity preservation, occlusion handling, and visual realism~\cite{li2020faceshifter,chen2020simswap,nirkin2019fsgan}. More recently, diffusion-based methods such as DiffFace introduce diffusion priors and facial guidance for face swapping~\cite{kim2022diffface}.

Another important family is face reenactment or motion transfer, where the target identity is preserved while facial dynamics, expression, pose, or mouth motion are driven by another source. Video-driven methods commonly encode facial motion through learned keypoints or deformation fields, as in the First Order Motion Model, Thin-Plate Spline Motion Model, and Face-vid2vid~\cite{siarohin2019fomm,zhao2022tpsmm, wang2021facevid2vid}. More recent portrait-animation methods, such as LivePortrait, further improve identity preservation and fine-grained motion control~\cite{guo2024liveportrait}. Audio-driven talking-face methods further manipulate lip and expression dynamics conditioned on speech~\cite{prajwal2020wav2lip}.

\subsection{Deepfake Detection}
Deepfake detection has traditionally relied on visual artifacts introduced during face synthesis and blending. Early CNN-based approaches detect mesoscopic patterns, warping distortions, texture defects, and blending boundaries in manipulated faces~\cite{afchar2018mesonet,rossler2019faceforensics,li2019warping,li2020facexray}. Although effective on established benchmarks, such low-level cues often generalize poorly as generation and post-processing techniques evolve.

To move beyond individual-frame artifacts, subsequent methods model temporal and behavioral inconsistencies, including abnormal eye blinking, head pose, lip motion, and frame-to-frame dynamics~\cite{li2018ictuoculi,yang2019headpose,haliassos2021lips,masi2020twobranch}. These cues provide more interpretable evidence by relating detection decisions to recognizable facial behaviors.

\subsection{Remote Photoplethysmography (rPPG)
for Deepfake Detection}
Physiology-aware deepfake detection has received increasing attention as physiological signals can offer a complementary source of semantically meaningful evidence by reflecting subtle temporal processes associated with human biological activity. FakeCatcher, DeepFakesON-Phys, and DeepRhythm argue that synthetic videos may fail to preserve coherent biological or heartbeat-related signals~\cite{ciftci2019fakecatcher,hernandez2020deepfakesonphys,qi2020deeprhythm}. More recent methods such as rPPGNet further combine rPPG priors with neural architectures for deepfake detection~\cite{dwivedi2026rppgnet}. These studies motivate the use of intrinsic physiological cues rather than only external visual artifacts. However, two gaps remain. First, many widely used deepfake datasets are not collected with physiological ground truth, making it difficult to verify whether extracted rPPG signals are physiologically meaningful. D'Amelio et al. partially addressed this issue by constructing UBFC1-F to investigate the effects of swap quality and video compression on remotely estimated cardiac signals~\cite{damelio2023using}. Nevertheless, their study is restricted to a small single-dataset setting and a single identity-swapping technique.

Second, existing methods often treat physiology as an independent feature stream or combine it with visual evidence through shallow fusion. In this paper, we address these limitations by constructing manipulations on real rPPG datasets and modeling bidirectional dependencies between rPPG tokens and facial behavior tokens. This enables the detector to jointly capture subtle pixel-level and frame-level variations at a deeper level, utilizing them as semantically interpretable cross-modal signals for deepfake detection.

\section{Methods}

\subsection{Aligned Physiology and Behavior Tokens}
Given a video clip of $T$ frames, we construct two temporally aligned token sequences: physiology tokens $R\in\mathbb{R}^{T\times d_r}$ and behavior tokens $B\in\mathbb{R}^{T\times d_b}$. In our experiments, $T=160$, $d_r=96$, and $d_b=26$. The physiology stream is extracted using a PhysFormer model trained for supervised rPPG estimation~\cite{yu2022physformer}. We spatially average its intermediate spatio-temporal features and interpolate them along the temporal dimension to obtain one 96-dimensional rPPG token per input frame. These intermediate representations are optimized for rPPG estimation and are therefore treated as pulse-sensitive rPPG representations, and the final model output is used as the estimated rPPG waveform for physiological information analysis. Thus, the tokens retain pulse-sensitive temporal features rather than reducing each clip to a single estimated heart rate.

For the behavior stream, OpenFace~\cite{baltrusaitis2018openface} is applied frame by frame. Each behavior token contains detection confidence, six head-pose parameters, two gaze angles, and 17 facial Action Unit (AU) measurements. The rPPG and OpenFace sequences are aligned using the same video identifier, clip start index, and temporal window. Frames for which OpenFace tracking fails are excluded through an attention mask.

\subsection{Bidirectional Physiology--Behavior Co-Attention}
The two input streams have different feature dimensions and semantics. We first map them into a shared $d$-dimensional space using separate learned projections followed by layer normalization:
\begin{equation}
R_0=\operatorname{LN}(RW_R+b_R),\qquad
B_0=\operatorname{LN}(BW_B+b_B).
\end{equation}
As shown in Figure~\ref{fig:overview}, we then model their cross-modal dependencies in both directions. The physiology-to-behavior branch uses $R_0$ as queries and $B_0$ as keys and values, whereas the behavior-to-physiology branch reverses these roles:
\begin{align}
C_R &= \operatorname{MHA}(Q=R_0,K=B_0,V=B_0),\\
C_B &= \operatorname{MHA}(Q=B_0,K=R_0,V=R_0).
\end{align}
Residual connections, dropout, and layer normalization produce the contextualized sequences
\begin{equation}
\widetilde R=\operatorname{LN}(R_0+\operatorname{Drop}(C_R)),\qquad
\widetilde B=\operatorname{LN}(B_0+\operatorname{Drop}(C_B)).
\end{equation}
Unlike concatenation or a global weighted sum, this operation allows each temporal token in one modality to retrieve complementary evidence from the other modality before temporal aggregation.

\subsection{Evaluation Protocol and Measures}

We use subject-disjoint 5-fold cross-validation. All rPPG identities are kept disjoint across folds, including both source and target identities involved in generated samples. For motion-transfer videos, driving sequences are also partitioned disjointly, ensuring that neither identities nor driving videos overlap between training and testing. Unless otherwise stated, results are reported as the mean $\pm$ standard deviation across folds. Physiological recovery is evaluated per clip using the processing protocol of rPPG-Toolbox~\cite{liu2022rppg}: differential PhysFormer outputs are integrated, detrended, and band-pass filtered between 0.6 and 3.3~Hz before the measures in Table~\ref{tab:evaluation_measures}(a) are computed. For the complementary one-dimensional separability analysis, we compare the recovered and reference signals using the waveform discrepancy descriptors in Table~\ref{tab:evaluation_measures}(b). 

\begin{table}[!h]
\centering
\caption{Domain-specific physiological measures and waveform discrepancy descriptors.}
\label{tab:evaluation_measures}
\scriptsize
\begin{subtable}[t]{\columnwidth}
\centering
\caption{Physiological recovery measures}
\begin{tabular}{@{}p{0.20\linewidth}p{0.73\linewidth}@{}}
\toprule
Measure & Definition \\
\midrule
HR MAE $\downarrow$ & Mean absolute difference, in bpm, between the dominant-frequency heart rates estimated from recovered and reference pulse signals. \\
HR RMSE $\downarrow$ & Square root of the mean squared heart-rate error, in bpm, which penalizes large estimation errors more strongly than MAE. \\
MACC $\uparrow$ & Maximum absolute Pearson cross-correlation between recovered and reference waveforms over all temporal lags. \\
SNR $\uparrow$ & Log power ratio, in dB, between recovered-signal energy within $\pm0.1$~Hz of the reference fundamental and second harmonic and the remaining energy in 0.6--3.3~Hz. \\
PSD Corr. $\uparrow$ & Pearson correlation between the z-normalized recovered and reference periodogram vectors within the 0.6--3.3~Hz physiological band. \\
\bottomrule
\end{tabular}
\end{subtable}%

\vspace{0.6em}
\begin{subtable}[t]{\columnwidth}
\centering
\caption{Waveform discrepancy descriptors}
\label{tab:waveform_descriptor_definitions}
\begin{tabular}{@{}p{0.20\linewidth}p{0.73\linewidth}@{}}
\toprule
Descriptor & Description \\
\midrule
Raw MAE & Mean absolute sample-wise difference between the z-normalized recovered and reference waveforms. \\
Raw RMSE & Root mean squared sample-wise difference between the z-normalized waveforms. \\
Raw Std. & Absolute difference between the standard deviations of the z-normalized waveforms. \\
Deriv. MAE & Mean absolute difference between the first temporal differences of the two waveforms. \\
Deriv. RMSE & Root mean squared difference between the first temporal differences. \\
Deriv. Std. & Absolute difference between the standard deviations of the first temporal differences. \\
\bottomrule
\end{tabular}
\end{subtable}
\end{table}

\section{Experiments}
\subsection{Deepfake generation}
\label{sec:dataset_construction}
Our experimental samples are generated from UBFC-rPPG and COHFACE recordings~\cite{bobbia2019ubfc,heusch2017cohface}, two facial-video datasets containing synchronized physiological measurements. We retain the selected original recordings as the real subset and construct two manipulated subsets using identity swapping and motion transfer. The detailed description of our dataset is listed in the table ~\ref{tab:dataset_overview}.

For identity swapping, subjects are paired separately within each dataset using a same-gender directed all-to-all protocol, excluding self-pairs, then after quality control, the videos are selected according to three-source-per-target protocol. For a source subject and a target recording, the source identity is transferred onto the target video while the target's facial motion, head pose, background, and temporal structure are preserved. We use the open-source identity-swapping framework FaceFusion ~\cite{facefusion2026} with its default model and SimSwap ~\cite{chen2020simswap}, to increase generator diversity. The generated source-target pairs are assigned approximately equally between the two frameworks to reduce the risk that the detector overfits to artifacts specific to a single identity-swapping implementation.

For motion transfer, we adapt the data-generation strategy proposed in Motion Matters~\cite{paruchuri2024motion}. We use Face-vid2vid and driving sequences from its associated TalkingHead-1KH dataset~\cite{wang2021facevid2vid} to animate the source rPPG recordings. Following the general driving-video selection protocol of Motion Matters, candidate sequences are characterized according to their head-pose and facial-action dynamics in order to cover diverse motion patterns. Each source subject is paired with three different driving videos after quality control.

% In the preamble

\begin{table}[t]
\centering
\caption{Overview of the deepfake manipulations applied to rPPG datasets.}
\label{tab:dataset_overview}

\begin{subtable}{\columnwidth}
\centering
\caption{Dataset composition after quality control.}
\label{tab:dataset_composition}
\resizebox{\columnwidth}{!}{
\begin{tabular}{lrrrrrr}
\toprule
Dataset &
Subjects &
Real &
Identity Swap &
Motion Transfer &
All Fake &
Total \\
\midrule
COHFACE~\cite{heusch2017cohface}
& 39 & 156 & 468 & 468 & 936 & 1,092 \\

UBFC-rPPG~\cite{bobbia2019ubfc}
& 42 & 42 & 126 & 126 & 252 & 294 \\
\midrule
Total
& 81 & 198 & 594 & 594 & 1,188 & 1,386 \\
\bottomrule
\end{tabular}
}
\end{subtable}

\vspace{0.8em}

\begin{subtable}{\columnwidth}
\centering
\caption{Duration, resolution, and frame rate of each subset.}
\label{tab:video_characteristics}
\resizebox{\columnwidth}{!}{
\begin{tabular}{llrrcc}
\toprule
Dataset &
Subset &
Videos &
Duration (h) &
Resolution &
FPS \\
\midrule
COHFACE
& Real
& 156 & 2.67
& $640{\times}480$ & 20 \\

COHFACE
& Identity swap
& 468 & 8.01
& $640{\times}480$ & 20 \\

COHFACE
& Motion transfer
& 468 & 7.96
& $640{\times}480$ & 20 \\
\midrule
UBFC-rPPG
& Real
& 42 & 0.22
& $640{\times}480$ & 30 \\

UBFC-rPPG
& Identity swap
& 126 & 0.65
& $640{\times}480$ & 30 \\

UBFC-rPPG
& Motion transfer
& 126 & 0.79
& $640{\times}480$ & 30 \\
\midrule
Total
& --
& 1,386 & 20.30
& -- & -- \\
\bottomrule
\end{tabular}
}
\end{subtable}

\end{table}

\subsection{Ground-truth Biological Signal Analysis on Real and Manipulated Videos}
\label{sec:physio_analysis}
A key advantage of our experimental setting is that the authentic videos are from rPPG datasets with synchronized physiological ground truth. This allows us to analyze biological-signal preservation using the contact-sensor pulse signals provided with the original rPPG recordings as an external reference. For motion-transfer videos, the external driving sequence provides facial motion only and is not treated as a physiological reference. Therefore, for all manipulated videos, we compare the estimated rPPG signal with the ground-truth pulse signal synchronized with the corresponding original rPPG source recording. In Table~\ref{tab:physio_recovery}, we report Heart rate (HR) estimation error, waveform agreement, and spectral consistency between the estimated rPPG signal and the provided ground-truth pulse signal. Lower HR MAE and RMSE indicate smaller heart-rate deviation, while higher MACC, SNR, and PSD correlation indicate stronger temporal and spectral biological consistency.

\begin{table}[t]
\centering
\caption{Biological signal consistency on real-synthetic video pairs.}
\label{tab:physio_recovery}
\resizebox{\columnwidth}{!}{
\begin{tabular}{lccccc}
\toprule
Type & HR MAE$\downarrow$ & HR RMSE$\downarrow$ & MACC$\uparrow$ & SNR$\uparrow$ & PSD Corr.$\uparrow$ \\
\midrule
Real & 3.13 & 6.95 & 0.753 & 2.31 & 0.820 \\
FaceSwap & 5.70 & 10.28 & 0.672 & 0.28 & 0.702 \\
MotionTrans. & 12.51 & 16.23 & 0.474 & -4.59 & 0.341 \\
\bottomrule
\end{tabular}
}
\end{table}

\begin{figure*}[t]
\centering
\includegraphics[width=0.92\linewidth]{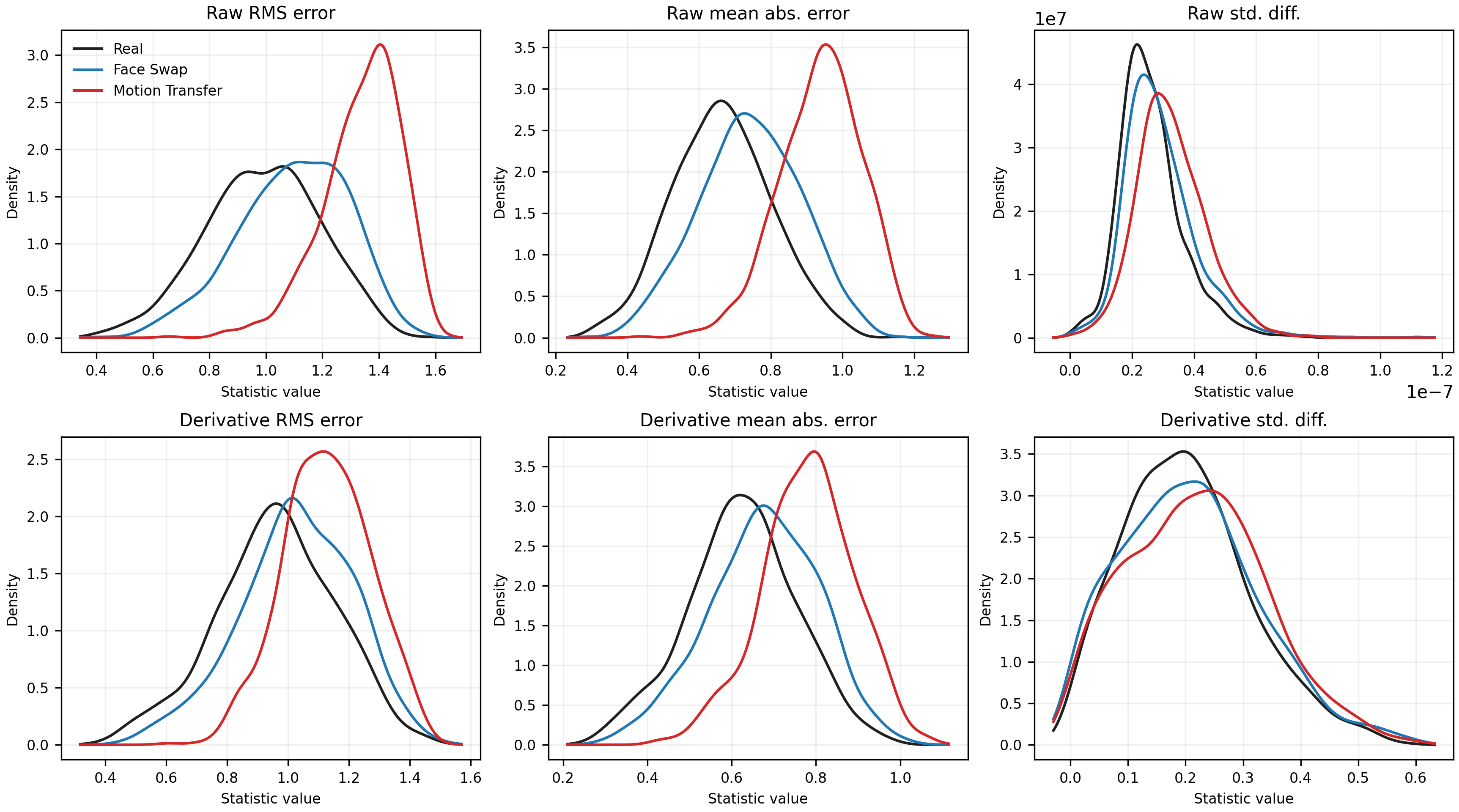}
\caption{Distribution of raw and derivative biological-signal discrepancy statistics. Each statistic is computed between the estimated rPPG signal and the dataset-provided ground-truth pulse.}
\label{fig:raw_deriv_stats}
\end{figure*}

As shown in Table~\ref{tab:physio_recovery}, real videos achieve the best physiological recovery across all metrics, with the lowest HR error and the highest waveform and spectral agreement, which confirms that the rPPG dynamics from authentic COHFACE and UBFC videos are strongly recovered.

Face Swap videos show a clear degradation, increasing HR MAE from 3.13 bpm to 5.70 bpm and reducing MACC from 0.753 to 0.672. However, the degradation is moderate: the PSD correlation remains relatively high at 0.702, suggesting that face wapping still preserves part of the target video's temporal physiological structure.

Motion Transfer videos exhibit the strongest physiological disruption. Compared with real videos, HR MAE increases by approximately four times, while MACC drops from 0.753 to 0.474 and PSD correlation drops from 0.820 to 0.341. The negative SNR further indicates that the recovered signal no longer concentrates energy around the ground-truth heart-rate frequency and its harmonic components.This suggests that motion transfer does not merely perturb the appearance of the face, but substantially alters the temporal signal structure used for rPPG recovery.

To further analyze the source of this degradation, we follow the spirit of pairwise biological-signal analysis and examine whether simple waveform and spectral statistics can separate authentic and manipulated clips. Figure~\ref{fig:raw_deriv_stats} visualizes the distributions of raw and derivative discrepancy statistics computed between the estimated rPPG signal and the dataset-provided ground-truth pulse. The distributions show a clear shift for Motion Transfer clips, especially in raw waveform discrepancy measures, indicating that motion-transfer synthesis introduces stronger temporal distortion of the recovered biological signal, which is consistent with the previous physiological recovery analysis.

\begin{table}[!t]
\centering
\caption{One-dimensional threshold separability using waveform discrepancy statistics.}
\label{tab:threshold_waveform_stats}
\scriptsize
\setlength{\tabcolsep}{2.5pt}
\resizebox{\columnwidth}{!}{
\begin{tabular}{llccc}
\toprule
Pair & Stat. & Th. & Acc. & AUC \\
\midrule
\multirow{6}{*}{Real vs. FaceSwap}
& Raw MAE & \textbf{0.711} & \textbf{62.00} & \textbf{0.655} \\
& Raw RMSE & 1.088 & 61.50 & 0.653 \\
& Raw Std. & 0.000 & 55.95 & 0.578 \\
& Deriv. MAE & 0.684 & 59.13 & 0.614 \\
& Deriv. RMSE & 0.984 & 58.33 & 0.602 \\
& Deriv. Std. & 0.225 & 52.69 & 0.510 \\
\midrule
\multirow{6}{*}{Real vs. MotionTrans.}
& Raw MAE & \textbf{0.826} & \textbf{86.16} & \textbf{0.934} \\
& Raw RMSE & 1.224 & 85.40 & 0.924 \\
& Raw Std. & 0.000 & 65.62 & 0.683 \\
& Deriv. MAE & 0.705 & 76.06 & 0.822 \\
& Deriv. RMSE & 1.028 & 69.42 & 0.755 \\
& Deriv. Std. & 0.287 & 58.37 & 0.545 \\
\midrule
\multirow{6}{*}{FaceSwap vs. MotionTrans.}
& Raw MAE & \textbf{0.887} & \textbf{77.98} & \textbf{0.860} \\
& Raw RMSE & 1.274 & 77.26 & 0.849 \\
& Raw Std. & 0.000 & 59.41 & 0.608 \\
& Deriv. MAE & 0.744 & 66.92 & 0.728 \\
& Deriv. RMSE & 1.041 & 61.32 & 0.659 \\
& Deriv. Std. & 0.001 & 56.74 & 0.534 \\
\bottomrule
\end{tabular}
}
\end{table}

\begin{table*}[!t]
\centering
\caption{Subject-disjoint 5-fold ablation of rPPG--behavior fusion under controlled deepfake manipulations applied to UBFC-rPPG and COHFACE. Results are video-level mean $\pm$ standard deviation across folds.}
\label{tab:rb_ablation}
\scriptsize
\setlength{\tabcolsep}{2.6pt}
\begin{subtable}[t]{0.495\textwidth}
\centering
\caption{Face swap}
\resizebox{\linewidth}{!}{%
\begin{tabular}{lcccc}
\toprule
Method & Acc. $\uparrow$ & F1 $\uparrow$ & AUC $\uparrow$ & EER $\downarrow$ \\
\midrule
Behavior only & 50.59 $\pm$ 1.18 & 29.32 $\pm$ 26.29 & 51.16 $\pm$ 1.09 & 49.38 $\pm$ 1.25 \\
rPPG only & 80.02 $\pm$ 2.41 & 79.65 $\pm$ 1.67 & 89.27 $\pm$ 4.34 & 18.11 $\pm$ 4.54 \\
Concat & 79.93 $\pm$ 4.21 & 79.17 $\pm$ 4.05 & 90.15 $\pm$ 4.27 & 17.55 $\pm$ 5.62 \\
Weighted sum & 79.52 $\pm$ 2.85 & 78.46 $\pm$ 3.99 & 88.61 $\pm$ 3.81 & 19.82 $\pm$ 5.85 \\
Bidirectional co-attn & \textbf{85.32 $\pm$ 7.66} & \textbf{85.19 $\pm$ 7.75} & \textbf{92.80 $\pm$ 5.65} & \textbf{14.75 $\pm$ 9.14} \\
\bottomrule
\end{tabular}%
}
\end{subtable}%
\hfill
\begin{subtable}[t]{0.495\textwidth}
\centering
\caption{Motion transfer}
\resizebox{\linewidth}{!}{%
\begin{tabular}{lcccc}
\toprule
Method & Acc. $\uparrow$ & F1 $\uparrow$ & AUC $\uparrow$ & EER $\downarrow$ \\
\midrule
Behavior only & 90.30 $\pm$ 7.04 & 85.44 $\pm$ 10.66 & 92.99 $\pm$ 3.55 & 15.51 $\pm$ 7.01 \\
rPPG only & 92.10 $\pm$ 1.10 & 91.31 $\pm$ 2.10 & 95.29 $\pm$ 2.36 & 8.25 $\pm$ 3.22 \\
Concat & 93.01 $\pm$ 1.92 & 91.77 $\pm$ 1.83 & 95.45 $\pm$ 3.21 & 3.01 $\pm$ 2.57 \\
Weighted sum & 93.57 $\pm$ 2.87 & 92.26 $\pm$ 1.48 & 96.12 $\pm$ 2.44 & 3.10 $\pm$ 1.79 \\
Bidirectional co-attn & \textbf{95.13 $\pm$ 1.42} & \textbf{94.99 $\pm$ 1.51} & \textbf{96.78 $\pm$ 2.83} & \textbf{2.20 $\pm$ 1.32} \\
\bottomrule
\end{tabular}%
}
\end{subtable}
\end{table*}

We further quantify this separability using a one-dimensional threshold analysis. For each scalar statistic, we search for the threshold that best separates two video categories and report the resulting accuracy and rank-based AUC. As shown in Table~\ref{tab:threshold_waveform_stats}, raw waveform discrepancies are the most discriminative, especially for Real vs. Motion Transfer and Face Swap vs. Motion Transfer. 

Therefore, these results provide preliminary evidence that videos manipulated from datasets with synchronized physiological ground truth can be distinguished using simple thresholding. Using only Raw MAE and a single optimized threshold, the Real--FaceSwap pairwise separation task achieves an initial accuracy of 62.00\% and an AUC of 0.655. This result is noteworthy given that it is obtained from only one scalar waveform descriptor. The Real--MotionTransfer task exhibits substantially stronger separability, reaching 86.16\% accuracy and 0.934 AUC. Furthermore, FaceSwap and MotionTransfer achieve 77.98\% pairwise accuracy and 0.860 AUC, indicating that different deepfake manipulation mechanisms degrade recoverable physiological information to different extents.

Overall, our experiment supports the hypothesis that deepfake generation degrades recoverable physiological signals, and that the degradation depends on the manipulation mechanism. The effect is visible not only in HR estimation error, but also in waveform alignment, spectral concentration, and PSD similarity to the ground-truth pulse. Because the reference signal is the dataset-provided physiological ground truth rather than another extracted facial ROI signal, the observed gap directly measures how much authentic physiological information is preserved or destroyed by each manipulation type. This signal-level evidence motivates the following detection experiments, where rPPG is modeled together with facial behavior rather than used as an isolated shallow cue.

\subsection{Physiology-Behavior Fusion Analysis}
\label{sec:fusion_analysis}

To evaluate whether physiological cues should be used as isolated descriptors or jointly modeled with facial behavior, we compare the proposed bidirectional co-attention with single-modality and existing fusion baselines. Prior physiology-aware detectors commonly exploit rPPG through handcrafted biological maps, independent physiological classifiers, temporal attention over heartbeat representations, or late integration with neural features~\cite{ciftci2019fakecatcher,hernandez2020deepfakesonphys,qi2020deeprhythm,dwivedi2026rppgnet}. Following this line of work, we include two simple fusion baselines: feature concatenation and learnable weighted summation. All methods use the same subject-disjoint 5-fold protocol and the same rPPG and OpenFace behavior tokens.

Table~\ref{tab:rb_ablation}(a) reports the results on face swapping, where coarse facial dynamics are largely inherited from the target video. Behavior alone is consequently close to chance, whereas rPPG alone reaches 89.27\% AUC, supporting the physiological-disruption analysis in the preceding section. Neither simple fusion strategy consistently improves upon this baseline: concatenation combines only temporally pooled descriptors, and weighted summation learns a global modality preference without resolving temporal cross-modal dependencies. Bidirectional co-attention instead exchanges token-level context before pooling and achieves the best result on every metric, including 85.32\% accuracy and 92.80\% AUC. This corresponds to AUC gains of 3.53 and 2.65 percentage points over rPPG alone and concatenation, respectively.

Motion transfer presents a different regime. As shown in Table~\ref{tab:rb_ablation}(b), behavior alone already achieves 90.30\% accuracy and 92.99\% AUC, which is likely due to the substantial modifications in expression and pose dynamics introduced by the externally driven animation. rPPG alone is also nearly perfect, consistent with the substantially larger physiological degradation measured in the experiment before. As the single modalities of behavior and rPPG both perform effectively, the task is therefore nearly saturated: bidirectional co-attention improves
the performance to 95.13\%, while the remaining fusion variants remain close to 93\%.

We further evaluate the applicability of the proposed representation on Celeb-DF-v2~\cite{li2020celebdf}. Unlike the deepfake samples generated for our experiments, Celeb-DF-v2 does not provide synchronized physiological ground truth; therefore, it is used only as a standard deepfake detection benchmark and is excluded from the physiological recovery analysis. The physiological and behavior extractors remain fixed, and their configurations remain identical to those used for the constructed manipulated data. The bidirectional co-attention detector is initialized from the checkpoint trained on our generated samples and subsequently fine-tuned on the Celeb-DF-v2 training split, rather than trained from scratch. During fine-tuning, we use class-weighted cross-entropy and balanced sampling to account for class imbalance.

Because Celeb-DF-v2 is substantially imbalanced, we use class-weighted cross-entropy together with balanced sampling during training. We adopt the best-performing detector configuration identified through controlled experiments involving deepfake manipulations applied to UBFC-rPPG and COHFACE, retaining the bidirectional co-attention architecture and its selected hyperparameters, and retrain it on Celeb-DF-v2 using class-weighted cross-entropy. The resulting model achieves 91.20\% video-level accuracy and 86.08\% AUC. The latter is lower than the 92.80\% face-swap AUC and 96.78\% motion-transfer AUC obtained on the manipulated samples. One possible explanation is the reduced reliability of rPPG representations extracted from Celeb-DF-v2, whose videos were not collected under conditions designed for physiological sensing. Despite the degradation in physiological feature quality, the bidirectional co-attention model retains a high raw accuracy, suggesting that the proposed co-attention fusion strategy has potential to generalize beyond physiology-oriented datasets.

\section{Conclusion and Future Work}

We studied deepfake detection from two complementary perspectives: the degradation of recoverable physiological signals and the disruption of their relationship with facial behavior. By applying face-swap and motion-transfer manipulations to COHFACE and UBFC-rPPG recordings, we retained synchronized physiological references and could directly quantify manipulation-dependent changes in heart-rate error, waveform agreement, and spectral consistency. Motion transfer caused the strongest degradation, while face swapping preserved more of the target's physiological structure. Consistent with this difference, the proposed bidirectional co-attention detector benefited from token-level interaction between PhysFormer and facial-behavior representations, achieving 92.80\% AUC on face swapping and 96.78\% AUC on motion transfer under subject-disjoint 5-fold evaluation. Its 86.08\% AUC after retraining on Celeb-DF-v2 shows promise outside controlled rPPG datasets, but does not yet establish strict cross-dataset robustness.

Future work should first characterize physiological degradation more comprehensively. Beyond the current HR, waveform-correlation, SNR, PSD-correlation, and discrepancy measures, this includes pulse-rate variability, phase coherence, spatial consistency across facial regions, harmonic structure, signal-quality indices, and uncertainty-aware estimates. Second, stricter generalization studies are needed on deepfake datasets, including train-on-one/test-on-another protocols, unseen generators and manipulation families, and systematic analysis of compression, frame rate, resolution, illumination, skin tone, motion, and occlusion. Third, the manipulation set should be expanded to newer swapping, reenactment, talking-face, and diffusion-based generators so that detector gains cannot be attributed to a small set of synthesis artifacts. Finally, reference-free physiological consistency objectives and modality-reliability gating may improve deployment on in-the-wild videos where contact-sensor ground truth is unavailable and one modality can be unreliable.

\section*{Ethical considerations and data availability.}
The manipulated videos were generated solely for deepfake-detection research using public UBFC-rPPG and COHFACE recordings. Authorization was obtained for the use of UBFC-rPPG, while COHFACE was used under its applicable research-access conditions. For data-protection reasons, the generated videos will not be released; code, configurations, data splits, and evaluation scripts will be made available upon publication, while reconstruction remains subject to the access conditions of the original datasets.

\bibliographystyle{ACM-Reference-Format}
\bibliography{references}
\end{document}